\documentclass[10pt,twocolumn,letterpaper]{article}

\usepackage[pagenumbers]{cvpr} 

\definecolor{cvprblue}{rgb}{0.21,0.49,0.74}
\usepackage[pagebackref,breaklinks,colorlinks,allcolors=cvprblue]{hyperref}
\usepackage{multirow}
\usepackage{wrapfig}
\usepackage{float}
\usepackage{cuted}
\usepackage[table]{xcolor}
\usepackage{algorithm}
\usepackage{algorithmic}
\usepackage{amsmath}
\usepackage{float}

\newcommand{\ourmethod}{{TopoMesh}}
\newcommand{\boldstartspace}[1]{\medskip\noindent\textbf{#1}}
\DeclareRobustCommand{\corresmark}{\textsuperscript{\textdagger}}

\usepackage{xcolor}

\definecolor{darkgreen}{RGB}{30,128,30}

\def\paperID{960} 
\def\confName{CVPR}
\def\confYear{2026}

\title{TopoMesh: High-Fidelity Mesh Autoencoding via Topological Unification}

\author{
    Guan Luo\textsuperscript{\rm 1,2}\quad
    Xiu Li\textsuperscript{\rm 2}\quad
    Rui Chen\textsuperscript{\rm 2,3}\quad
    Xuanyu Yi\textsuperscript{\rm 2}\quad
    Jing Lin\textsuperscript{\rm 2}\\
    Chia-Hao Chen\textsuperscript{\rm 1}\quad
    Jiahang Liu\textsuperscript{\rm 2}\quad
    Song-Hai Zhang\textsuperscript{\rm 1}\thanks{Corresponding authors.}\quad
    Jianfeng Zhang\textsuperscript{\rm 2}\corresmark\\
    \textsuperscript{\rm 1} Tsinghua University\quad
    \textsuperscript{\rm 2} ByteDance Seed\quad
    \textsuperscript{\rm 3} HKUST\\
    \url{https://logan0601.github.io/projects/topomesh/index.html}
}

\begin{document}

\maketitle
\begin{strip}
    \vspace{-22.5mm}
	\centering
	\includegraphics[width=1\textwidth]{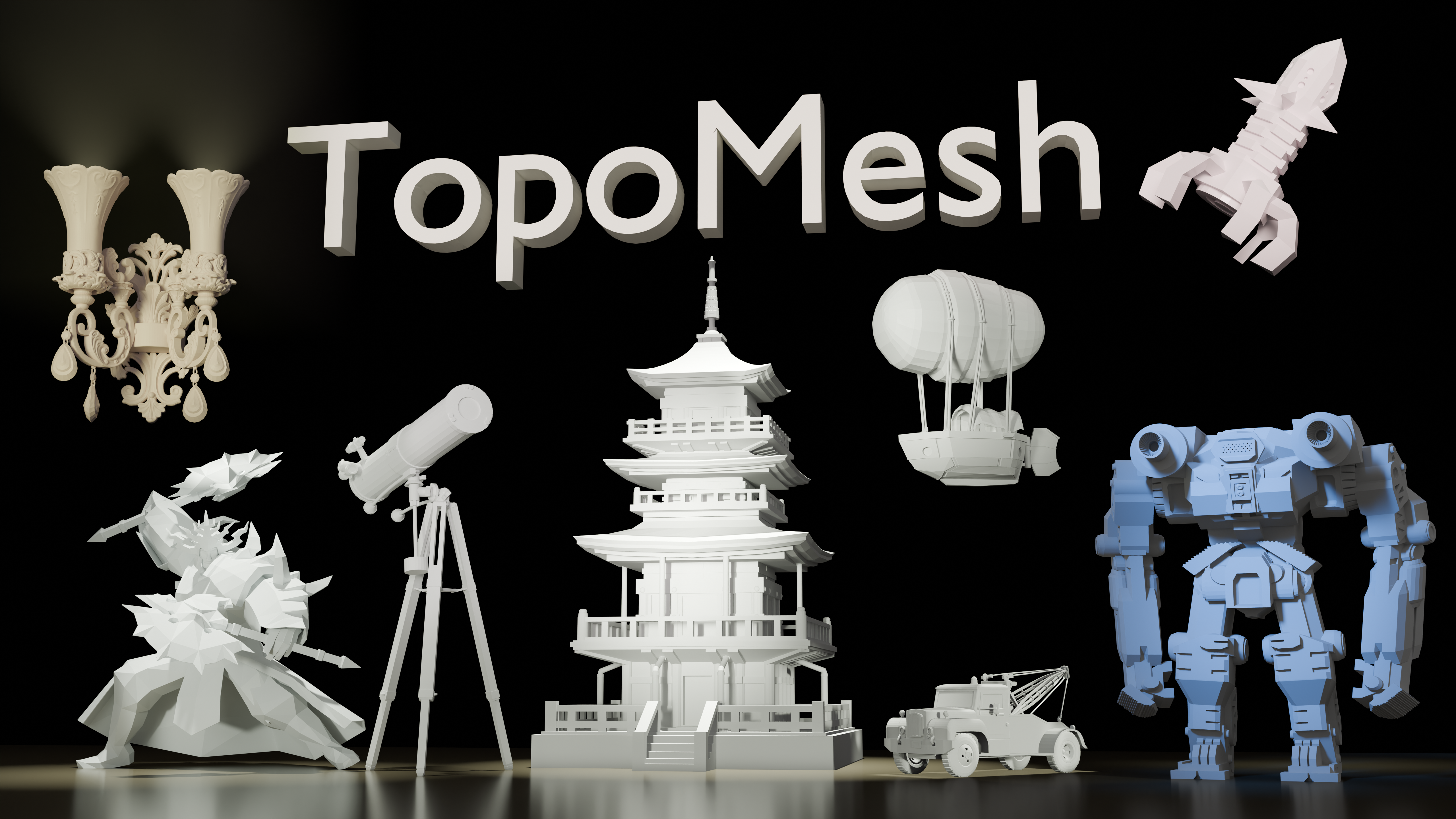}
    \vspace{-7.5mm}
    \captionof{figure}{High-fidelity mesh at $1024^3$ resolution reconstructed by TopoMesh within 5s.}
    \label{fig:teaser}
    \vspace{-4mm}
\end{strip}

\begin{abstract}

The dominant paradigm for high-fidelity 3D generation relies on a VAE-Diffusion pipeline, where the VAE's reconstruction capability sets a firm upper bound on generation quality. A fundamental challenge limiting existing VAEs is the \textit{representation mismatch} between ground-truth meshes and network predictions: GT meshes have arbitrary, variable topology, while VAEs typically predict fixed-structure implicit fields (\eg, SDF on regular grids).
This inherent misalignment prevents establishing explicit mesh-level correspondences, forcing prior work to rely on indirect supervision signals such as SDF or rendering losses. Consequently, fine geometric details, particularly sharp features, are poorly preserved during reconstruction.
To address this, we introduce \ourmethod{}, a sparse voxel-based VAE that unifies both GT and predicted meshes under a shared Dual Marching Cubes (DMC) topological framework. Specifically, we convert arbitrary input meshes into DMC-compliant representations via a remeshing algorithm that preserves sharp edges using an L$\infty$ distance metric. Our decoder outputs meshes in the same DMC format, ensuring that both predicted and target meshes share identical topological structures. 
This establishes explicit correspondences at the vertex and face level, allowing us to derive explicit mesh-level supervision signals for topology, vertex positions, and face orientations with clear gradients.
Our sparse VAE architecture employs this unified framework and is 
trained with Teacher Forcing and progressive resolution training 
for stable and efficient convergence.
Extensive experiments demonstrate that \ourmethod{} significantly outperforms existing VAEs in reconstruction fidelity, achieving superior preservation of sharp features and geometric details, as shown in \cref{fig:teaser}.

\end{abstract}
\section{Introduction}
\label{sec:intro}

Deep generative models are rapidly transforming 3D content creation, with applications spanning gaming, virtual reality, robotics, and computer-aided design. Among various approaches, 3D native diffusion models~\cite{3dshape2vecset, trellis} have emerged as a leading paradigm, attributed to their superior generation quality, strong generalization, and scalability. These models operate in a latent space, requiring a powerful Variational Autoencoder (VAE)~\cite{vae} to compress irregular, topologically-varying meshes into regular latent representations and reconstruct them reliably. Consequently, the VAE's reconstruction fidelity acts as the primary bottleneck, setting a firm upper bound on the generation quality.

A fundamental challenge limiting existing VAEs is the \textit{representation mismatch} between ground-truth meshes and network predictions. Ground-truth meshes exhibit arbitrary topology, such as irregular connectivity and varying vertex counts, while VAEs typically predict fixed-structure representations. This inherent structural gap prevents establishing explicit mesh-level correspondence, forcing previous methods to rely on indirect supervision signals, such as SDF or rendered images, which introduce distinct limitations. Early methods like 3DShape2VecSet~\cite{3dshape2vecset}, Clay~\cite{clay}, and TripoSG~\cite{triposg} predict implicit SDF and extract meshes via Marching Cubes (MC)~\cite{mc}, which constrain vertices to lie on grid edges, inevitably smoothing sharp edges and corners. Recent works like Trellis~\cite{trellis} and SparseFlex~\cite{sparseflex} replace MC with the more expressive FlexiCubes~\cite{flexicubes} decoder but pivot to rendering-based supervision, {which introduces supervisory ambiguity.} For example, gradients for fine geometric details are often lost due to limited resolution, occlusion, and sparse viewpoints. Other methods like Direct3D-S2~\cite{direct3ds2} and Sparc3D~\cite{sparc3d} return to SDF supervision and MC-based extraction, thus still suffering from the grid-alignment constraints. Consequently, a critical question remains: 
\textbf{How can we design a VAE that possesses both the expressive power to faithfully represent sharp features and the structural alignment necessary to enable direct, unambiguous supervision on mesh?}

To answer this question, we introduce \ourmethod{}, a novel framework that resolves the representation mismatch through Topological Unification. By ensuring both network predictions and ground-truth geometry share a unified Dual Marching Cubes (DMC)~\cite{dmc} framework, our method establishes explicit correspondence at the vertex and face level. This enables, for the first time, direct supervision on mesh topology, vertex positions, and face orientations.

To realize topological unification on both ground-truth geometry and network outputs, \ourmethod{} introduces two core components. First, Topo-Remesh, a robust, fully GPU-accelerated algorithm designed to convert arbitrary input meshes into feature-preserving, DMC-compliant representations. Unlike traditional methods~\cite{clay, dora} that use L2 distance, a point-to-point metric that rounds sharp corners, we introduce an L$\infty$ distance that incorporates local surface structure to preserve sharp features. Combined with a Manifold Dual Contouring extractor~\cite{odc}, Topo-Remesh produces high-fidelity watertight outputs at $1024^3$ resolution in approximately 15 seconds.
Second, Topo-VAE operates on the unified DMC topology with a sparse encoder and a decoupled decoder.
Our encoder employs sparse voxel-point cross-attention mechanism, where each voxel attends only to points within it, enabling efficient processing of millions of points. Our decoder builds upon FlexiCubes but decouples the parameters into topology and geometry components. Combined with the DMC framework that establishes correspondences, this enables direct supervision on topology, positions, and orientations.
These two components synergistically enable high-fidelity mesh autoencoding with direct supervision.

With the unified architecture in place, a principled training strategy is essential for realizing the full potential of our VAE. 
In preliminary experiments, we observed severe instability when applying geometry losses only upon correct topology prediction. The root cause is a tug-of-war: when topology becomes correct, its loss vanishes while large geometry losses suddenly activate, introducing gradients that often flip the topology back to incorrect and prevent convergence.
To break this cycle, we introduce Teacher Forcing: 
during training, we provide the decoder 
with ground-truth topology, allowing geometry components to learn under stable, correct topological configurations. 
At test time, the decoder independently predicts both.
The strategy is complemented by ground-truth guided voxel pruning and progressive resolution training to accelerate overall convergence.

Our main contributions are: \textbf{(1)} A novel paradigm for VAE-based mesh autoencoding that unifies predictions and ground truth under a shared DMC topological framework, enabling direct supervision on fundamental mesh attributes (topology, positions, orientations). \textbf{(2)} A robust, fully GPU-accelerated remeshing algorithm that converts arbitrary input meshes into DMC-compliant representations while preserving sharp geometric features. \textbf{(3)} A sparse voxel-based VAE with efficient voxel-point cross-attention encoding and a FlexiCubes-based decoder with topology-geometry decoupling. \textbf{(4)} A comprehensive training strategy with Teacher Forcing, voxel pruning, and progressive resolution training. \textbf{(5)} Extensive experiments demonstrating state-of-the-art reconstruction fidelity, achieving 8\% improvement in F-Score and superior sharp feature preservation compared to existing methods.

\begin{figure*}[ht]
    \centering
    \vspace{-2mm}
    \includegraphics[width=0.99\linewidth]{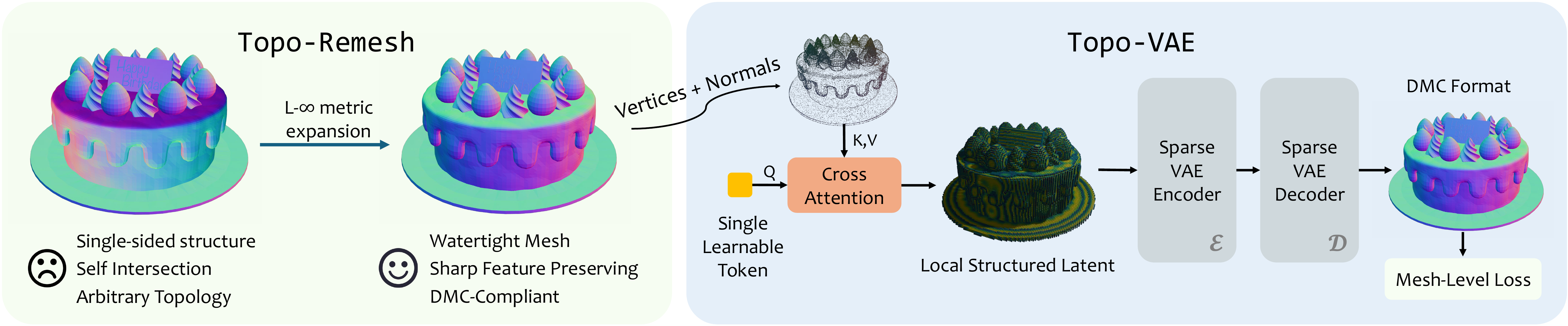}
    \vspace{-2mm}
    \caption{\ourmethod{} comprises two core modules. \textbf{Topo-Remesh} converts imperfect, real-world meshes into clean, DMC-compliant representations while preserving sharp features. \textbf{Topo-VAE} takes vertices with normals as input and reconstructs the mesh in the same DMC format. With topological unification, we apply direct supervision on mesh attributes: topology, vertex positions, and face orientations.}
    \label{fig:method}
    \vspace{-4mm}
\end{figure*}

\section{Related Work}
\label{sec:relatedwork}

\boldstartspace{3D Generation}. Existing 3D generation methods roughly fall into three categories. DreamFusion~\cite{dreamfusion} and subsequent works~\cite{magic3d, fantasia3d, zero123, prolificdreamer, dreamgaussian, sjc, latentnerf, richdreamer} leverage a pre-trained 2D diffusion model to optimize the radiance fields~\cite{nerf, 3dgs} via Score Distillation Sampling (SDS). Multi-view generation~\cite{mvdream, syncdreamer, wonder3d, zero123pp, sv3d} followed by sparse-view reconstruction~\cite{lrm, instant3d, gslrm, lgm, instantmesh, crm, tgs, meshlrm, grm, meshformer} forms a fast pipeline, but is limited in geometric fidelity and generation stability. 3D native diffusion models~\cite{3dshape2vecset, clay, trellis, triposg, hunyuan3d, sparseflex, sparc3d, direct3d, direct3ds2, 3dtopiaxl, xcube, one2345pp, octfusion, ultra3d} have emerged as a leading paradigm for their high quality and strong generalization. Crucially, these models operate in the latent space, requiring a powerful VAE for high-fidelity mesh reconstruction.

\boldstartspace{3D Shape VAEs}. Modern 3D VAEs can be broadly categorized by latent representation. VecSet-based VAEs~\cite{3dshape2vecset, clay, triposg, dora, craftsman, michelangelo, hunyuan3d, lagem} represent shapes as a global set of latent vectors. While effective at capturing overall shape, their global nature limits the modeling of fine-grained details.
Sparse Voxel-based VAEs~\cite{trellis, sparseflex, direct3ds2, sparc3d, xcube, hi3dgen, unilat}, which our work builds upon, have emerged as the state-of-the-art for high-resolution modeling. These methods leverage sparse voxel grids to represent complex shapes. However, they are caught in a difficult trade-off regarding supervision. On one hand, methods like Trellis~\cite{trellis} and SparseFlex~\cite{sparseflex} employ rendering-based supervision on an expressive FlexiCubes~\cite{flexicubes} decoder, thus suffer from ambiguous gradients for fine geometric details. On the other hand, methods like Direct3D-S2~\cite{direct3ds2} and Sparc3D~\cite{sparc3d} revert to SDF supervision but use MC-based decoders, re-encountering sharp feature loss during the extraction stage.

\boldstartspace{Isosurface Extraction}. Marching Cubes~\cite{mc} generates vertices via linear interpolation on grid edges, which inherently prevents the representation of sharp features. To overcome this, Dual Contouring~\cite{dc} and Dual Marching Cubes~\cite{dmc} generate a topologically dual mesh with vertices freely placed inside each grid cell, dramatically enhancing expressive power for sharp features. This dual paradigm has been made fully differentiable for end-to-end learning in FlexiCubes~\cite{flexicubes} and adapted for occupancy fields in ODC~\cite{odc}. Neural Methods such as NMC~\cite{nmc} and NDC~\cite{ndc} replace analytical rules with neural networks that directly predict vertex positions, but suffer from either complex parameterizations or a lack of manifold guarantees.

\section{Method}
\label{sec:method}

In this section, we present \ourmethod{} for high-fidelity mesh autoencoding, as demonstrated in \cref{fig:method}. We first introduce the architecture of \textit{Topo-VAE} (\cref{sec:vae}), including sparse voxel-point encoding and decoupled decoding, which outputs meshes under DMC format. We then describe the explicit mesh-level loss (\cref{sec:loss}) and training strategy (\cref{sec:train}) that realize the full potential of our VAE. Finally, we introduce \textit{Topo-Remesh} (\cref{sec:remesh}), a robust algorithm that converts any mesh into a clean, DMC-compliant representation while preserving sharp features.

\subsection{Topo-VAE}
\label{sec:vae}

The central objective of Topo-VAE is efficient encoding and differentiable decoding of mesh attributes, \ie, vertex placement and connectivity. This allows us to directly apply supervision on these fundamental attributes, which we argue is key to preserving fine geometric details.
Specifically, we encode an input mesh, represented by its vertices $V_i$ and its normals $N_i$, into compact, sparse voxels, and decode them back into an explicit mesh defined by output vertices $V_o$ and faces $F_o$. This process can be succinctly formulated as:
\begin{equation}
z = \mathcal{E}(V_i, N_i), \quad (V_o, F_o) = \mathcal{D}(z)
\end{equation}
where $\mathcal{E}$ and $\mathcal{D}$ represent encoder and decoder, respectively. We detail the core design of these components below.

\begin{figure}
    \centering
    \vspace{-2mm}
    \includegraphics[width=0.99\linewidth]{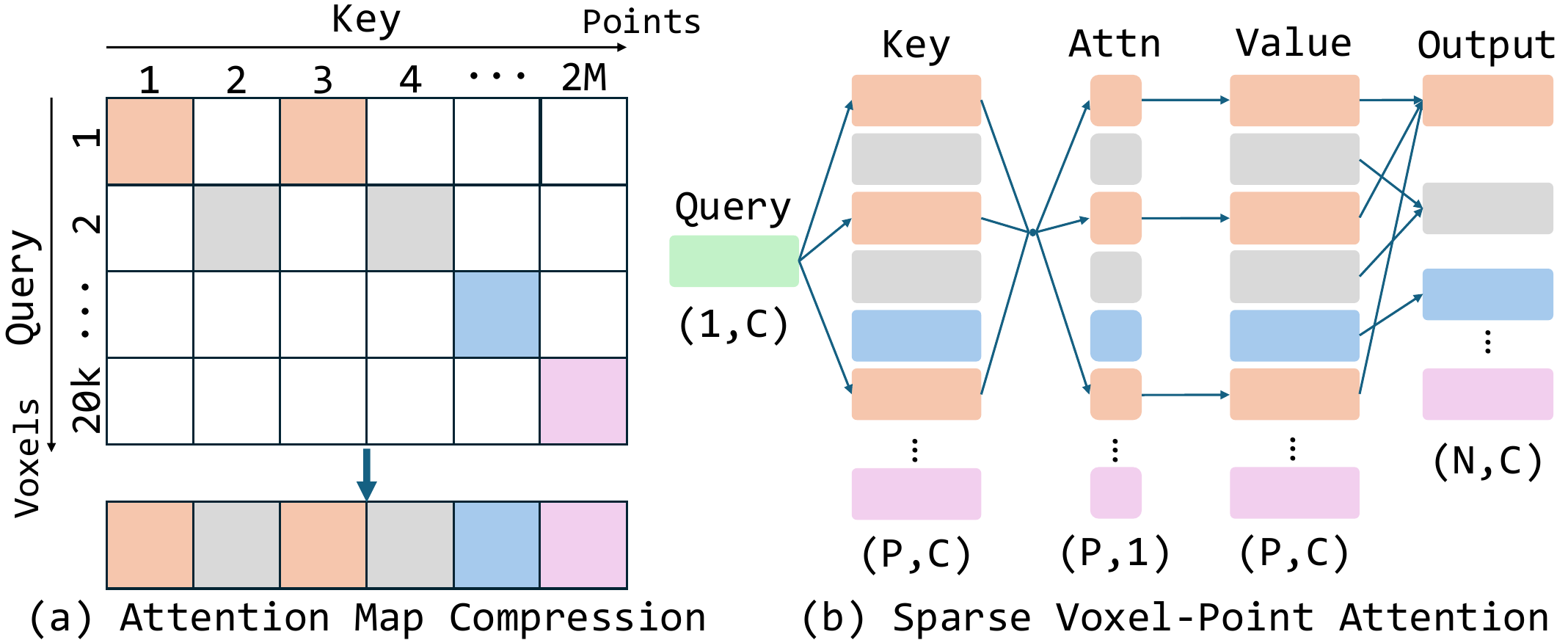}
    \vspace{-2mm}
    \caption{(a) Compression of the sparse attention map. (b) A single, shared query aggregates all keys from the same voxel to form the output feature of this voxel.}
    \vspace{-4mm}
    
    \label{fig:encoder}
\end{figure}

\boldstartspace{Sparse Voxel-Point Encoder}.
We treat input mesh vertices as a point cloud and encode them into sparse voxel features.
Naively encoding dense points into sparse voxels is computationally intractable. For example, attending from $20k$ sparse voxels at $64^3$ resolution to 2M 
input points requires an attention map of 74GB. Our key observation is that each point lies exclusively within a single voxel, enabling us to replace full attention with sparse local attention where each point interacts only with its enclosing voxel. As illustrated in \cref{fig:encoder}a, this local mechanism compresses the attention map: each column contains only one non-zero entry, reducing storage from $O(N\times P)$ to $O(P)$ (74GB to 3.8MB in our example), where $N$ and $P$ denote the number of voxels and points, respectively.
We further reduce computation by normalizing point coordinates within each voxel to local coordinates, which allows all voxels to share a single learnable query token, as demonstrated in \cref{fig:encoder}b. 
The sparse voxel-point cross-attention is formulated as follows:
\begin{equation}
    O_i=\sum_{j=1}^{n_i}\text{Softmax}_i(\frac{QK_j^T}{\sqrt{d}})\cdot V_j
\end{equation}
where $O_i$ is the output feature of voxel $i$, computed by aggregating features of $n_i$ points within it. The query $Q$ is the shared learnable token, and $K_j$ and $V_j$ are linear projections of the $j$-th point features within voxel $i$. The Softmax normalizes attention 
scores over all points within the voxel.

\boldstartspace{Decoupled Topology and Geometry Decoder}. Our decoder builds upon FlexiCubes~\cite{flexicubes}, which parameterizes mesh generation through: SDF values $s$ and deformation vectors $\delta$ for voxel corners, and interpolation weights $\alpha$, $\beta$, $\gamma$ for voxels controlling the precise vertex placement.
However, the coupled nature of SDF in standard FlexiCubes creates training instability: the sign of $s$ determines topology, while its magnitude affects geometry, resulting in entangled supervision signals that compete during optimization.

We redesign this process to enable direct mesh-level supervision and stable training. Specifically, we decouple the SDF $s$ into occupancy $o$ (sign) and magnitude $u$, categorizing parameters into topology and geometry components:
\begin{equation}
    \text{Topo} = \{o,\gamma\},\quad \text{Geom}=\{u,\alpha, \beta, \delta\}
\end{equation}
This separation allows mesh generation to proceed in two stages: output faces $F_o$ are determined exclusively by topological parameters, while vertices $V_o$ depend on all parameters:
\begin{equation}
    F_o=\text{DMC}(o)
\end{equation}
\begin{equation}
    V_o=\text{FlexiCubes}(o\times u, \alpha,\beta,\delta,\gamma)\label{eq:vert}
\end{equation}
where $\text{DMC}$ extracts mesh connectivity from the Dual Marching Cubes framework, and $\text{FlexiCubes}$ interpolates vertex positions differentiably. This explicit separation is crucial for stable training: by independently supervising topology (via occupancy) and geometry (via vertex placements), we prevent the ``tug-of-war" instability that arises when entangled topology and geometry losses compete during optimization (detailed in \cref{sec:train}).

\subsection{Explicit Mesh-Level Loss}
\label{sec:loss}

\begin{figure}
    \centering
    \vspace{-2mm}
    \includegraphics[width=0.99\linewidth]{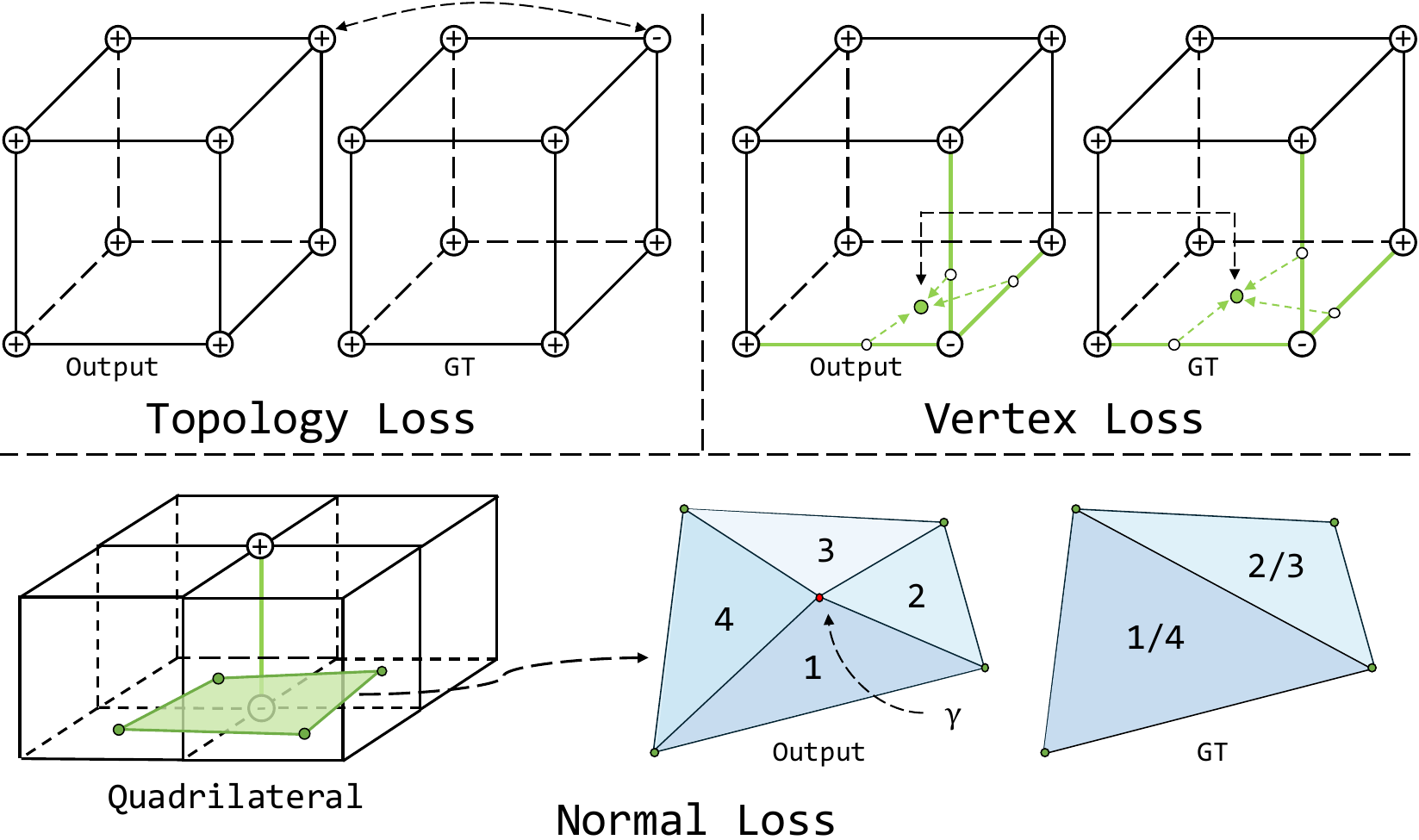}
    \vspace{-2mm}
    \caption{Illustration of explicit mesh-level loss.}
    \vspace{-4mm}
    \label{fig:loss}
\end{figure}

Our topological unification framework enables direct supervision on mesh attributes, \ie, topology, vertex positions, and surface orientations, as illustrated in Fig.~\ref{fig:loss}. Rather than supervising intermediate representations such as SDF values, our losses operate directly on the output mesh. Topology is supervised via occupancy $o$ at grid corners, while vertex and normal losses drive the optimization of all geometric parameters ($u$, $\alpha$, $\beta$, $\delta$) to produce accurate geometry.

\boldstartspace{Topology Loss}. We supervise occupancy $o$ at grid corners with Binary Cross-Entropy (BCE) Loss against the ground-truth signs $o_\text{gt}$:
\begin{equation}
    \mathcal{L}_\text{topo}=\text{BCE}(o, o_\text{gt})
\end{equation}

\boldstartspace{Vertex Loss}. 
To supervise vertex placement, we apply L1 loss directly on vertex positions $v$ against ground-truth positions $v_\text{gt}$. 
\begin{equation}
    \mathcal{L}_\text{vert}=\text{L1}(v,v_\text{gt})
\end{equation}

\boldstartspace{Normal Loss}. 
To supervise surface orientation and quad triangulation, we apply L1 loss on face normals $n$.
DMC produces quadrilaterals, which FlexiCubes triangulates using the $\gamma$ parameters.
However, FlexiCubes employs different triangulation strategies: splitting each quad into four triangles during training but only two during inference.
To provide supervision for the finer triangulation, we duplicate each ground-truth triangle to supervise its corresponding pair of predicted triangles, as illustrated in Fig.~\ref{fig:loss}:
\begin{equation}
    \mathcal{L}_\text{normal}=\text{L1}(n, n_\text{gt})
\end{equation}

\boldstartspace{Total Loss}. In addition to mesh-level losses, we include regularization terms: the FlexiCubes regularization loss $L_\text{reg}$~\cite{flexicubes}, a consistent loss $L_\text{con}$ that penalizes variance of attributes on grid corners~\cite{trellis}, voxel pruning loss $L_\text{occ}$ during upsampling~\cite{sparseflex}, and the standard KL-divergence loss $L_\text{KL}$~\cite{vae}. The final loss is a weighted combination:
\begin{align*}
    L=&\lambda_\text{topo}\mathcal{L}_\text{topo}+\lambda_\text{vert}\mathcal{L}_\text{vert}+\lambda_\text{normal}\mathcal{L}_\text{normal}\\
    &+\lambda_\text{occ}\mathcal{L}_\text{occ}+\lambda_\text{con}\mathcal{L}_\text{con}+\lambda_\text{reg}\mathcal{L}_\text{reg}+\lambda_\text{KL}\mathcal{L}_\text{KL}
\end{align*}

\begin{figure}
    \centering
    \vspace{-3mm}
    \includegraphics[width=0.99\linewidth]{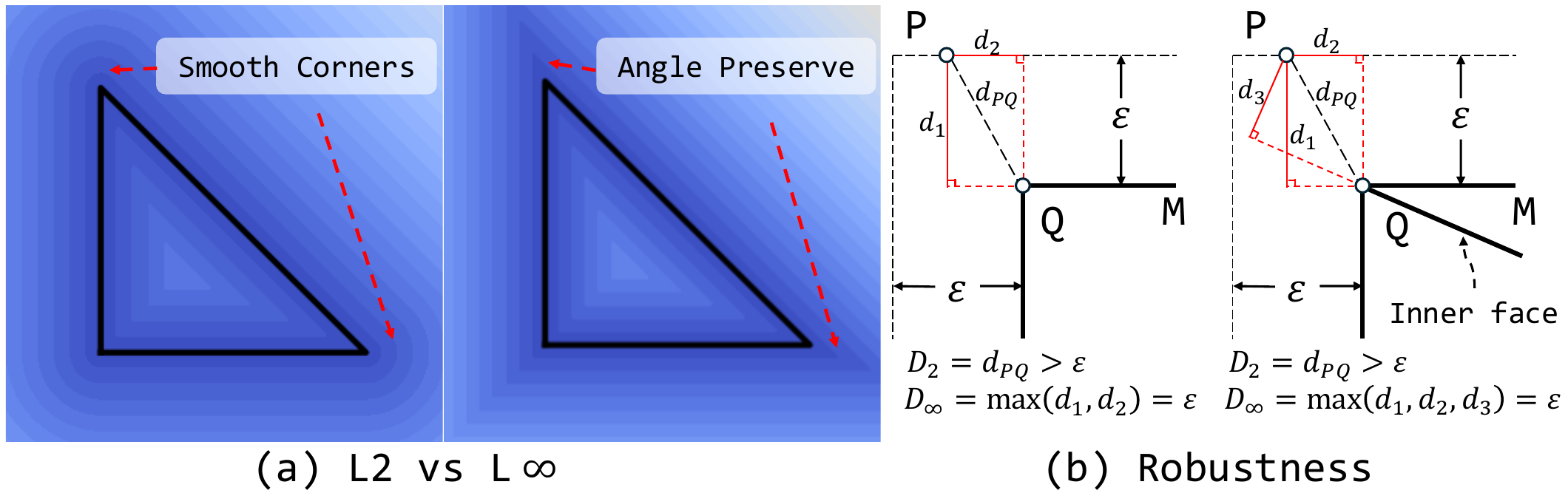}
    \vspace{-2mm}
    \caption{(a) Angle-preserving of L$\infty$ metric. (b) 2D illustration of L$\infty$ distance. $P$ lies on the isosurface that preserves angles.}
    \vspace{-4mm}
    \label{fig:sdf}
\end{figure}

\subsection{Training Scheme}
\label{sec:train}

Our SDF decoupling design (\cref{sec:vae}) provides the foundation for stable training by separating topology and geometry into independent parameter sets. 
However, we empirically find this insufficient to prevent training instability.
We trace the remaining challenges to a ``tug-of-war" dynamic between topology and geometry supervision. Initially, geometry losses provide no meaningful signal as the network struggles with topology. Once the network begins predicting correct topology for a region, geometry losses suddenly activate with large magnitudes, introducing disruptive gradients that destabilize topology learning. This creates an oscillation where neither topology nor geometry converges stably.
To break this cycle, we introduce several training strategies that leverage our decoupled architecture to stabilize and accelerate convergence.

\boldstartspace{Teacher Forcing}. 
To bypass the conditional dependency between topology and geometry, we employ Teacher Forcing during training. Instead of using the network's predicted occupancy in Eq.~\ref{eq:vert}, we provide ground-truth topology $o_\text{gt}$ to the decoder.
This ensures that geometric parameters receive stable, meaningful gradients from the first iteration, as they operate under correct topological configurations. While this introduces a discrepancy between training and inference, we observe a negligible impact on reconstruction quality, which we attribute to the smoothness of our latent space and the decoder's ability to predict accurate topology after sufficient training.

\boldstartspace{GT-guided Voxel Pruning}. 
Voxel pruning is essential for training efficiency at high resolutions. However, pruning based on early network predictions can be detrimental, often removing critical voxels and creating holes~\cite{sparc3d,sparseflex} and training instability~\cite{xcube}. We adopt a GT-guided strategy that preserves sparse voxels within a narrow band around the ground-truth surface during training, ensuring both geometric integrity and computational efficiency.

\boldstartspace{Progressive Resolution Training}. 
Our local-centric encoder design is adaptive to multiple resolutions, enabling progressive training. 
We begin at coarse resolution, allowing the model to capture global shape, then progress to finer resolutions where the model refines geometric details. This coarse-to-fine curriculum accelerates overall training convergence.

\subsection{Topo-Remesh}
\label{sec:remesh}

\begin{figure}
    \centering
    \vspace{-4mm}
    \includegraphics[width=0.99\linewidth]{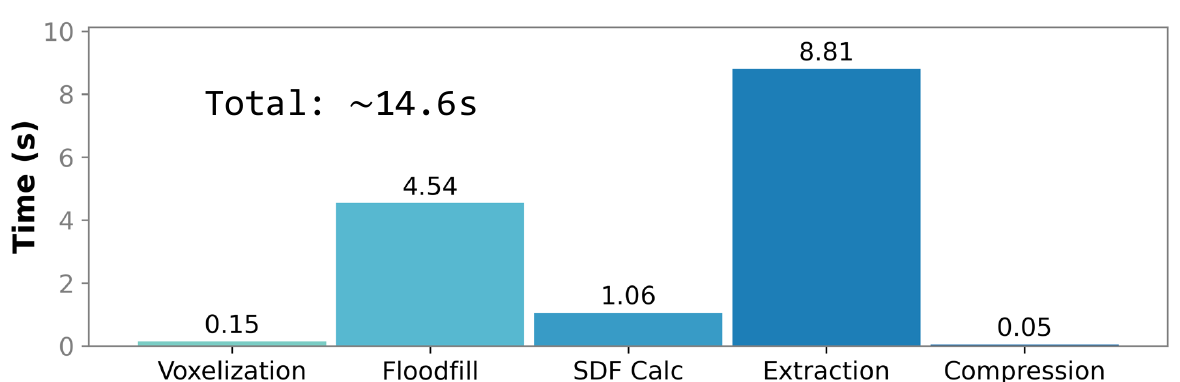}
    \vspace{-2mm}
    \caption{Remesh pipeline and execution time.}
    \vspace{-4mm}
    \label{fig:time}
\end{figure}

\begin{figure*}
    \centering
    \vspace{-4mm}
    \includegraphics[width=0.9\linewidth]{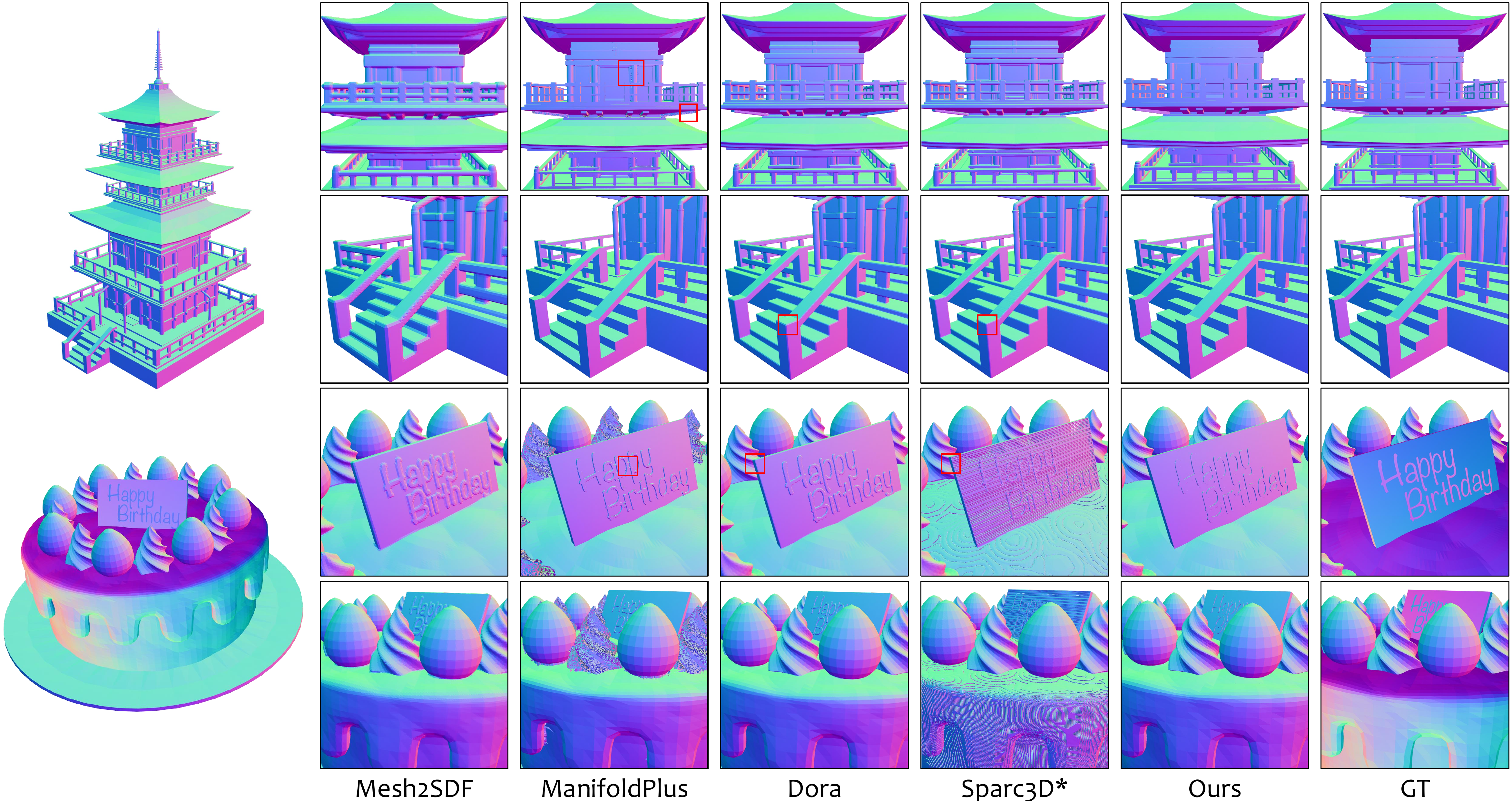}
    \vspace{-3mm}
    \caption{Visual comparison of remeshing. Our method produces clean meshes with sharp features. $*$ means a re-implementation version.}
    \label{fig:remesh}
\end{figure*}
\begin{table*}[htbp]
    \centering
    \vspace{-3mm}
    \caption{Quantitative result of remeshing. Our method preserves fine details with the highest efficiency. CD is multiplied by $10^5$.}
    \label{tab:remesh}
    \vspace{-3mm}
    \scalebox{0.9}{
    \begin{tabular}[with=0.99\linewidth]{r|c|ccc|ccc|cc}
    \toprule
    \multirow{2}[1]{*}{Method} & \multirow{2}[1]{*}{Device} & \multicolumn{3}{c|}{Thingi10K~\cite{thingverse}} & \multicolumn{3}{c|}{Objaverse~\cite{objaverse}} & \multicolumn{2}{c}{Efficiency} \\
    & & CD \(\downarrow\) & F1 \(\uparrow\) & ANC \(\uparrow\) & CD \(\downarrow\) & F1 \(\uparrow\) & ANC \(\uparrow\) & Time \(\downarrow\) & Size \(\downarrow\) \\ 
        \midrule
        Mesh2SDF~\cite{mesh2sdf} & CPU & 1.980 & 0.932 & 0.955 & 3.349 & 0.931 & 0.829 & 552.3 & \cellcolor{orange!25}66.5 \\
        ManifoldPlus~\cite{manifoldplus} & CPU & \cellcolor{red!60}\textbf{1.347} & 0.976 & \cellcolor{orange!25}0.981 & \cellcolor{red!60}\textbf{0.620} & \cellcolor{red!60}\textbf{0.991} & 0.780 & \cellcolor{orange!25}79.4 & 92.7 \\
        Dora~\cite{dora} & GPU & 1.492 & 0.972 & 0.970 & 1.057 & 0.987 & \cellcolor{orange!25}0.961 & 116.3 & 112.6 \\
        Sparc3D~\cite{sparc3d} & GPU & \cellcolor{orange!25}1.436 & \cellcolor{red!60}\textbf{0.978} & 0.975 & 2.864 & 0.970 & 0.929 & 175.9 & 121.6 \\
        Ours & GPU & 1.479 & \cellcolor{orange!25}0.978 & \cellcolor{red!60}\textbf{0.984} & \cellcolor{orange!25}0.964 & \cellcolor{orange!25}0.988 & \cellcolor{red!60}\textbf{0.964} & \cellcolor{red!60}\textbf{18.5} & \cellcolor{red!60}\textbf{28.7} \\
        \bottomrule
    \end{tabular}
    }
    \vspace{-5mm}
\end{table*}


Our approach demands that ground-truth meshes share the same DMC structure as our decoder to enable the direct mesh-level correspondences (Sec.~\ref{sec:vae},~\ref{sec:loss}).
To meet this requirement, we develop Topo-Remesh, a remeshing pipeline that converts arbitrary meshes into DMC-compliant representations while preserving geometric fidelity. 
A fundamental challenge in this process is surface dilation, a necessary step to ensure the output encloses a proper volume.
Prior methods~\cite{clay, dora} rely on L2-based dilation, which inherently smooths sharp corners, as illustrated in \cref{fig:sdf}a. While post-processing methods~\cite{pamo, sparc3d, manifoldplus} have been proposed to recover sharp features, they often introduce artifacts such as self-intersections and surface noise.

\boldstartspace{L$\infty$ Metric}. 
The limitation of L2 distance stems from its point-wise nature. As shown in \cref{fig:sdf}b, for a spatial point $P$ and its nearest point $Q$ on mesh $M$, standard $L_2$ distance measures only the direct distance between $P$ and $Q$, ignoring the local surface structure around $Q$ and thus rounding corners during dilation.
To preserve angles, we need a distance formulation that incorporates local geometry. Inspired by properties of the L$\infty$ norm~\cite{Linf}, we define the distance from point $P$ by considering all triangles $T(Q)$ incident to $Q$. Our L$\infty$ distance is the maximum L2 distance from $P$ to the plane of each incident triangle:
\begin{equation}
    D_\infty(P,Q)=\max_{T_i\in T(Q)}d(P,\Pi_i)
    \label{eq:linf}
\end{equation}
where $\Pi_i$ is the plane containing triangle $T_i$, and $d(P,\Pi_i)$ denotes the Euclidean distance from point $P$ to plane $\Pi_i$. The formulation is inherently angle-preserving: inflating the local surface at $Q$ by offsetting each incident plane outward by $\varepsilon$ creates a polyhedral envelope, and $P$ lies on its boundary.
This formulation is robust to topological defects and incorrect normals, as it requires no knowledge of topology or face directions, and is also robust to internal structure.
Please refer to the supplementary for more details.

\boldstartspace{Fully GPU-accelerated Remeshing}. 
As shown in Fig.~\ref{fig:time}, our pipeline consists of five sequential stages: Voxelization, Flood-fill, SDF Calculation, Isosurface Extraction, and Compression. We first voxelize the input mesh, then use flood-fill to locate voxels within a narrow band around the surface boundary. We compute L$\infty$ distance at grid corners to further identify surface-intersection voxels.
For surface extraction, we employ ODC~\cite{odc}, which extracts meshes that preserve sharp features through iteratively querying occupancy on grid edges and faces to compute the optimal vertex positions.
All stages are implemented on GPU, enabling end-to-end remeshing at $1024^3$ resolution in approximately 15 seconds.

\boldstartspace{DMC-based Mesh Compression}. We design an efficient compression scheme for high-resolution DMC-based mesh structures.
Specifically, for meshes up to $1024^3$ resolution, we store a set of primitives, including integer coordinates of valid voxels ($3\times10$ bits), occupancy of voxel corners ($8$ bits), intra-voxel vertex offsets ($3\times10$ bits), and triangulation decisions ($3$ bits), which can be quickly reassembled to form a mesh.
This representation achieves a $76\%$ compression ratio, competitive with Draco~\cite{draco}'s 84\%, while being orders of magnitude faster at 0.05 seconds versus Draco's 7 seconds for encoding and decoding.

\section{Experiment}
\label{sec:experiment}

We evaluate our method on both remeshing and autoencoding tasks. We first 
describe the experimental setup (\cref{sec:setting}), then present results for 
Topo-Remesh (\cref{sec:exp_remesh}) and Topo-VAE (\cref{sec:exp_vae}), followed 
by ablation studies (\cref{sec:ablation}). Finally, we present image-to-3D generation results using Topo-VAE (\cref{sec:generation}).

\begin{figure*}
    \centering
    \vspace{-4mm}
    \includegraphics[width=0.91\linewidth]{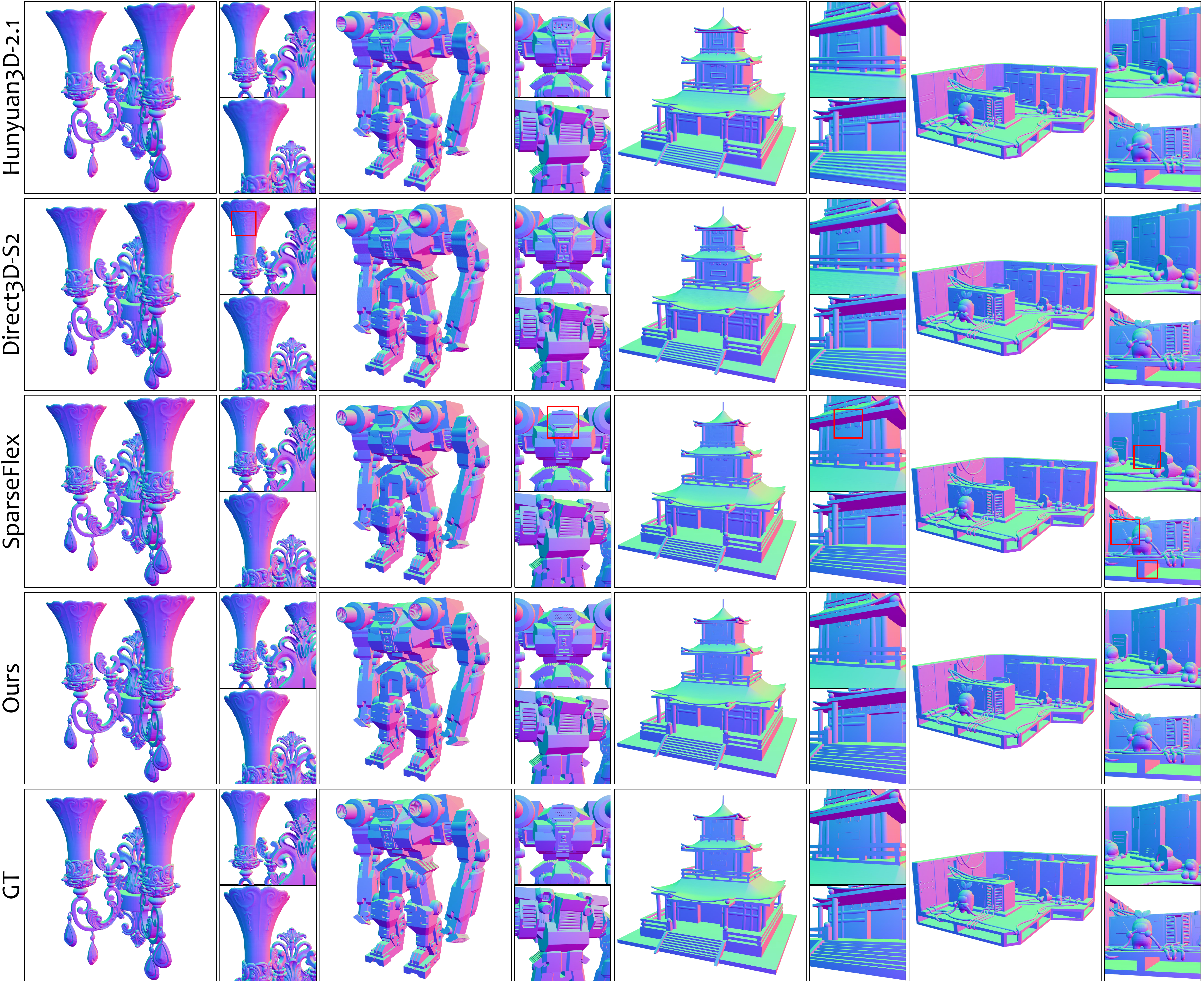}
    \vspace{-4mm}
    \caption{Visual comparison of VAE reconstruction. Our method better preserves 
    sharp features and fine details.}
    \label{fig:vae}
    \vspace{-4mm}
\end{figure*}

\subsection{Setting}
\label{sec:setting}
\boldstartspace{Dataset}. 
The training dataset contains $320k$ high-quality meshes from Sketchfab~\cite{objaverse}. For evaluating remeshing, we sample 500 objects each from Objaverse~\cite{objaverse} and Thingi10K~\cite{thingverse} (1k total). For autoencoding evaluation on complex geometries, we use the L3-L4 subset of Dora-Bench~\cite{dora}, which comprises 1.4k objects from GSO~\cite{gso}, ABO~\cite{abo}, Meta~\cite{meta}, and Objaverse~\cite{objaverse}. We further introduce \textit{Topo-Bench}, a supplementary benchmark of 1k objects from Objaverse~\cite{objaverse}, Thingi10K~\cite{thingverse}, and Toys4K~\cite{toys4k}, selected based on their number of sharp edges.

\boldstartspace{Training}. 
Training is conducted using AdamW~\cite{adamw} with a constant learning rate of $0.0001$. We employ the progressive resolution schedule for both VAE and DiT. The VAE is trained with a batch size of 64 for 160k, 160k, and 380k steps at $32^3$, $64^3$, and $128^3$ resolutions, respectively (700k steps total), and DiT is trained with a batch size of 512 for 200k, 300k, and 300k steps at $32^3$, $64^3$, $128^3$ resolutions. (800k steps total).

\subsection{Remesh}
\label{sec:exp_remesh}

\boldstartspace{Baselines}. 
We compare against Mesh2SDF~\cite{mesh2sdf}, ManifoldPlus~\cite{manifoldplus}, Dora~\cite{dora}, and a re-implemented version of Sparc3D~\cite{sparc3d}. We use Chamfer Distance (CD), F1 Score (F1), and Absolute Normal Consistency (ANC) as metrics, with all outputs reconstructed at $1024^3$ resolution.

\boldstartspace{Quality}. 
We present both qualitative and quantitative comparisons in Fig.~\ref{fig:remesh} and Tab.~\ref{tab:remesh}. All L$2$-based methods, including Mesh2SDF, Dora, and Sparc3D,  fail to preserve sharp edges and corners, as highlighted by red boxes. While projection-based (ManifoldPlus) and rendering-based (Sparc3D) refinement can slightly improve 3D metrics, they often introduce artifacts, such as self-intersections and noise. 
In contrast, our method produces clean meshes that faithfully preserve sharp features, such as the railings and handrails of the attic and birthday cards.

\boldstartspace{Efficiency}. 
Tab.~\ref{tab:remesh} shows that our method processes $1024^3$-resolution meshes in 18.5 seconds, significantly faster than baselines, which require at least one minute. Our compression yields compact outputs, averaging 28.7MB per mesh.

\subsection{Autoencoding}
\label{sec:exp_vae}

\begin{table*}[htbp]
    \centering
    \vspace{-2mm}
    \caption{Quantitative result of VAE. All methods except Trellis produce meshes at $1024^3$ resolution. Our method outperforms other methods in preserving sharp features. CD is multiplied by $10^5$.}
    \label{tab:vae}
    \vspace{-3mm}
    \scalebox{0.9}{
    \begin{tabular}[with=0.99\linewidth]{r|cc|cccc|cccc}
    \toprule
    \multirow{2}[1]{*}{Method} & \multirow{2}[1]{*}{\#Latent} & \multirow{2}[1]{*}{\#Dim} & \multicolumn{4}{c|}{Topo-Bench} & \multicolumn{4}{c}{Dora-Bench~\cite{dora}} \\
    & & & CD \(\downarrow\) & F1 \(\uparrow\) & F1-S \(\uparrow\) & ANC \(\uparrow\) & CD \(\downarrow\) & F1 \(\uparrow\) & F1-S \(\uparrow\) & ANC \(\uparrow\)  \\ 
        \midrule
        TripoSG~\cite{triposg} & 4096 & 64 & 2.658 & 0.893 & 0.715 & 0.965 & 1.697 & 0.959 & 0.717 & 0.976 \\
        Dora~\cite{dora} & 4096 & 64 & 2.167 & 0.905 & 0.754 & 0.968 & 1.814 & 0.964 & 0.768 & 0.977 \\
        Hunyuan3D-2.1~\cite{hunyuan3d} & 4096 & 64 & 2.538 & 0.888 & 0.767 & 0.965 & \cellcolor{orange!25}1.606 & 0.954 & 0.770 & 0.980 \\
        Trellis~\cite{trellis} & 12832 & 8 & 18.616 & 0.583 & 0.308 & 0.893 & 14.716 & 0.715 & 0.279 & 0.915 \\
        Direct3D-S2~\cite{direct3ds2} & 76386 & 16 & 2.713 & 0.813 & 0.694 & 0.920 & 2.313 & 0.881 & 0.819 & 0.968 \\
        SparseFlex~\cite{sparseflex} & 244691 & 8 & \cellcolor{red!60}1.840 & \cellcolor{red!60}0.920 & \cellcolor{orange!25}0.873 & \cellcolor{orange!25}0.992 & 1.625 & \cellcolor{red!60}0.973 & \cellcolor{orange!25}0.844 & \cellcolor{orange!25}0.994 \\
        Ours & 56006 & 8 & \cellcolor{orange!25}1.882 & \cellcolor{orange!25}0.917 & \cellcolor{red!60}0.932 & \cellcolor{red!60}0.993 & \cellcolor{red!60}1.126 & \cellcolor{orange!25}0.973 & \cellcolor{red!60}0.915 & \cellcolor{red!60}0.995 \\
        \bottomrule
    \end{tabular}
    }
    \vspace{-2mm}
\end{table*}

\boldstartspace{Baselines}. 
We compare against VecSet-based VAEs including TripoSG~\cite{triposg}, Dora~\cite{dora}, Hunyuan3D-2.1~\cite{hunyuan3d}; and Sparse Voxel-based VAEs including Trellis~\cite{trellis}, Direct3D-S2~\cite{direct3ds2}, SparseFlex~\cite{sparseflex}. Beyond standard metrics (CD, F1, ANC), 
we introduce F1-Sharp to quantify sharp feature preservation: an F1 Score computed on points sampled near edges and corners with dihedral angles sharper than 30 degrees, using the sharp edge sampling method from Dora~\cite{dora}. 
All methods generate meshes at $1024^3$ resolution, except Trellis ($256^3$).

\boldstartspace{Reconstruction Quality} 
Fig.~\ref{fig:vae} and Tab.~\ref{tab:vae} show that our method better reconstructs sharp corners (90-degree turns on sharp corners) and fine geometric details (holes on the robot), whereas all baselines fail to recover.
While SparseFlex achieves comparable metrics on overall shapes, our method uses only a quarter of its tokens and demonstrates significantly better sharp feature preservation (5.9\% and 7.1\% improvement in F1-Sharp on the two benchmarks).

\subsection{Ablation Study}
\label{sec:ablation}

\begin{figure}
    \centering
    \vspace{-3mm}
    \includegraphics[width=0.99\linewidth]{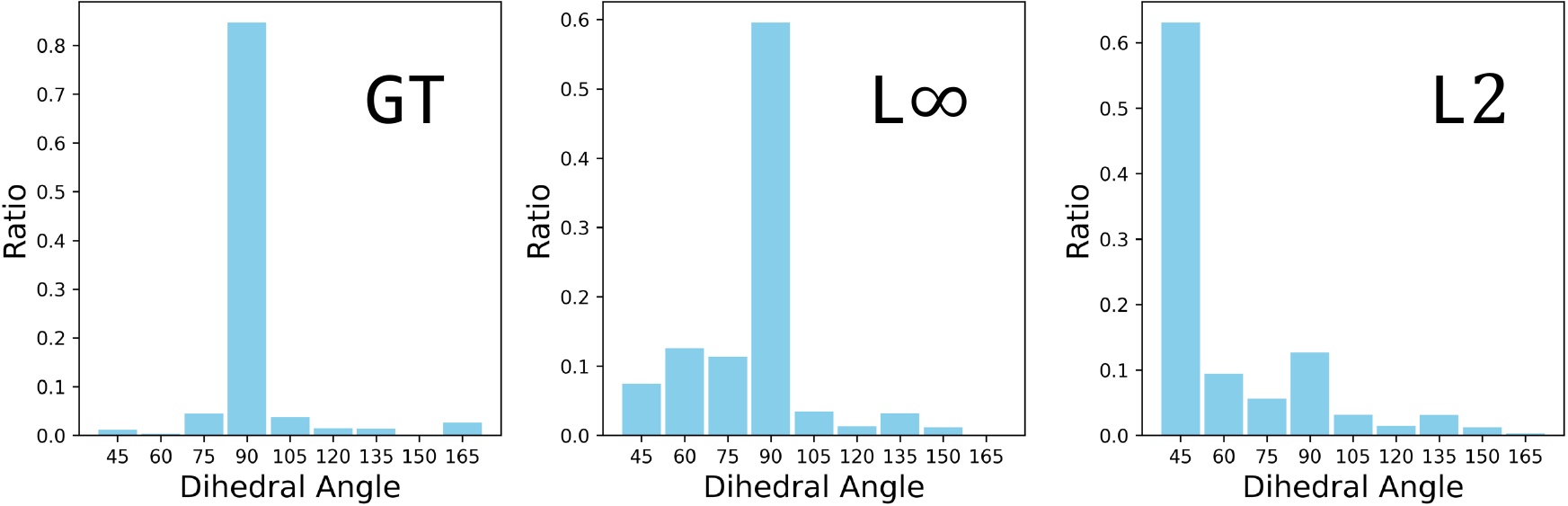}
    \vspace{-3mm}
    \caption{Dihedral angle distributions from different remeshing methods.}
    \label{fig:ablate_linf}
    \vspace{-1mm}
\end{figure}

\boldstartspace{L$\infty$ Distance}. We replace  L$\infty$ with L$2$ distance in our remeshing pipeline and visualize the dihedral angle distribution in \cref{fig:ablate_linf}. The histogram shows that our L$\infty$ distance preserves the distribution of sharp angles, whereas the L2 variant smooths the geometry, collapsing sharp features into nearly planar surfaces.

\begin{figure}
    \centering
    \vspace{-2mm}
    \includegraphics[width=0.9\linewidth]{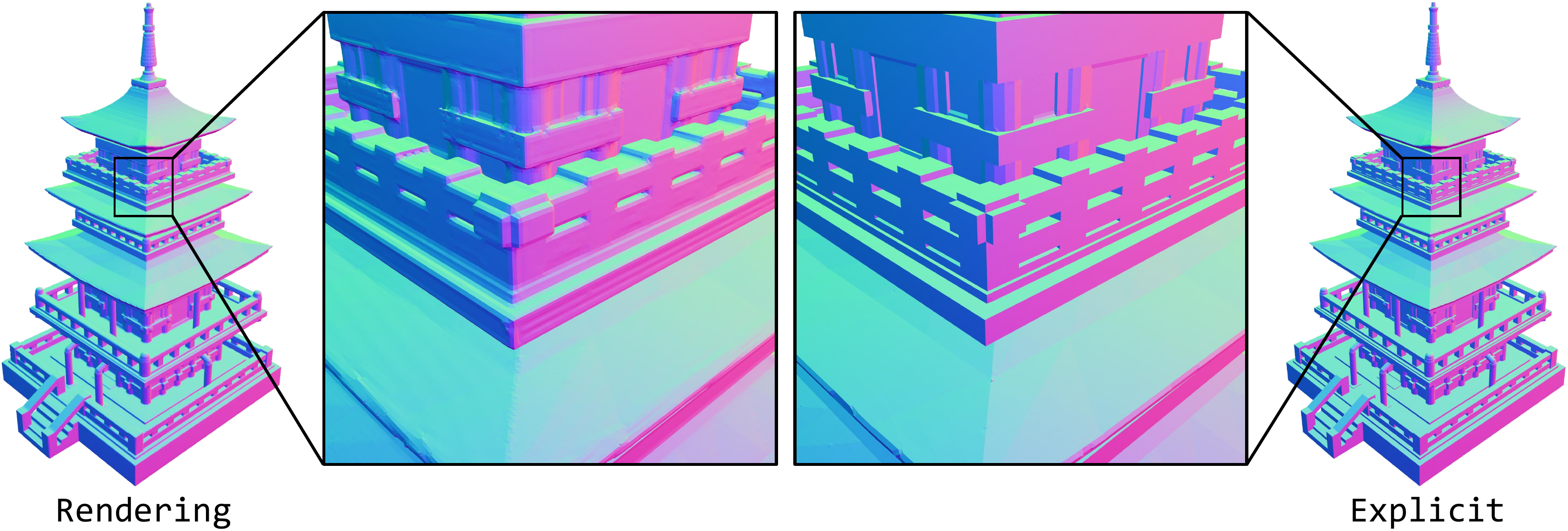}
    \vspace{-3mm}
    \caption{Comparison of direct mesh-level supervision versus rendering-based supervision.}
    \label{fig:ablate_loss}
    \vspace{-1mm}
\end{figure}

\begin{table}
    \centering
    \vspace{-2mm}
    \caption{Quantitative results of ablations.}
    \label{tab:ablate}
    \vspace{-2mm}
    \scalebox{0.9}{
    \begin{tabular}[with=0.99\linewidth]{r|cccc}
    \toprule
    Method & CD \(\downarrow\) & F1 \(\uparrow\) & F1-S \(\uparrow\) & ANC \(\uparrow\) \\ 
        \midrule
        render loss & 1.731 & 0.776 & 0.711 & 0.932 \\
        mesh-level loss & 0.150 & 0.975 & 0.991 & 0.999 \\
        \hline
        32res & 1.812 & 0.933 & 0.790 & 0.968 \\
        64res & 1.693 & 0.940 & 0.869 & 0.983 \\
        128res & 1.126 & 0.973 & 0.915 & 0.995 \\
        \bottomrule
    \end{tabular}
    }
    \vspace{-4mm}
\end{table}

\boldstartspace{Mesh-Level Supervision} 
We conduct a single-shape overfitting experiment to compare direct mesh-level 
supervision against rendering-based supervision. As shown in \cref{fig:ablate_loss} 
and \cref{tab:ablate} (first two rows), rendering-based supervision cannot 
reconstruct fine details and sharp edges accurately. In contrast, direct 
supervision enables nearly lossless reconstruction, demonstrating the advantage 
of our mesh-level correspondence framework.

\boldstartspace{Multi-Resolution Inference}. Our VAE is adaptive to varying resolutions. We show quantitative results on Dora-bench in ~\cref{tab:ablate} (Last three rows), each compared with its corresponding ground truth.

\begin{table}
    \centering
    \vspace{-3mm}
    \caption{Quantitative generation results on Toys4K.}
    \label{tab:generation}
    \vspace{-3mm}
    \scalebox{0.67}{
    \begin{tabular}[with=0.99\linewidth]{r|ccc|c}
    \toprule
    Method & Hunyuan3D-2.1~\cite{hunyuan3d} & Trellis~\cite{trellis} & Direct3D-S2~\cite{direct3ds2} & Ours \\ 
        \midrule
        FID \(\downarrow\) & 59.43 & 59.61 & 45.33 & \textbf{42.48} \\
        KID ($\times10^3$) \(\downarrow\) & 5.97 & 6.03 & 5.47 & \textbf{4.63} \\
        \bottomrule
    \end{tabular}
    }
    \vspace{-2mm}
\end{table}

\subsection{3D Generation}
\label{sec:generation}

\begin{figure}
    \centering
    \vspace{-1mm}
    \includegraphics[width=0.93\linewidth]{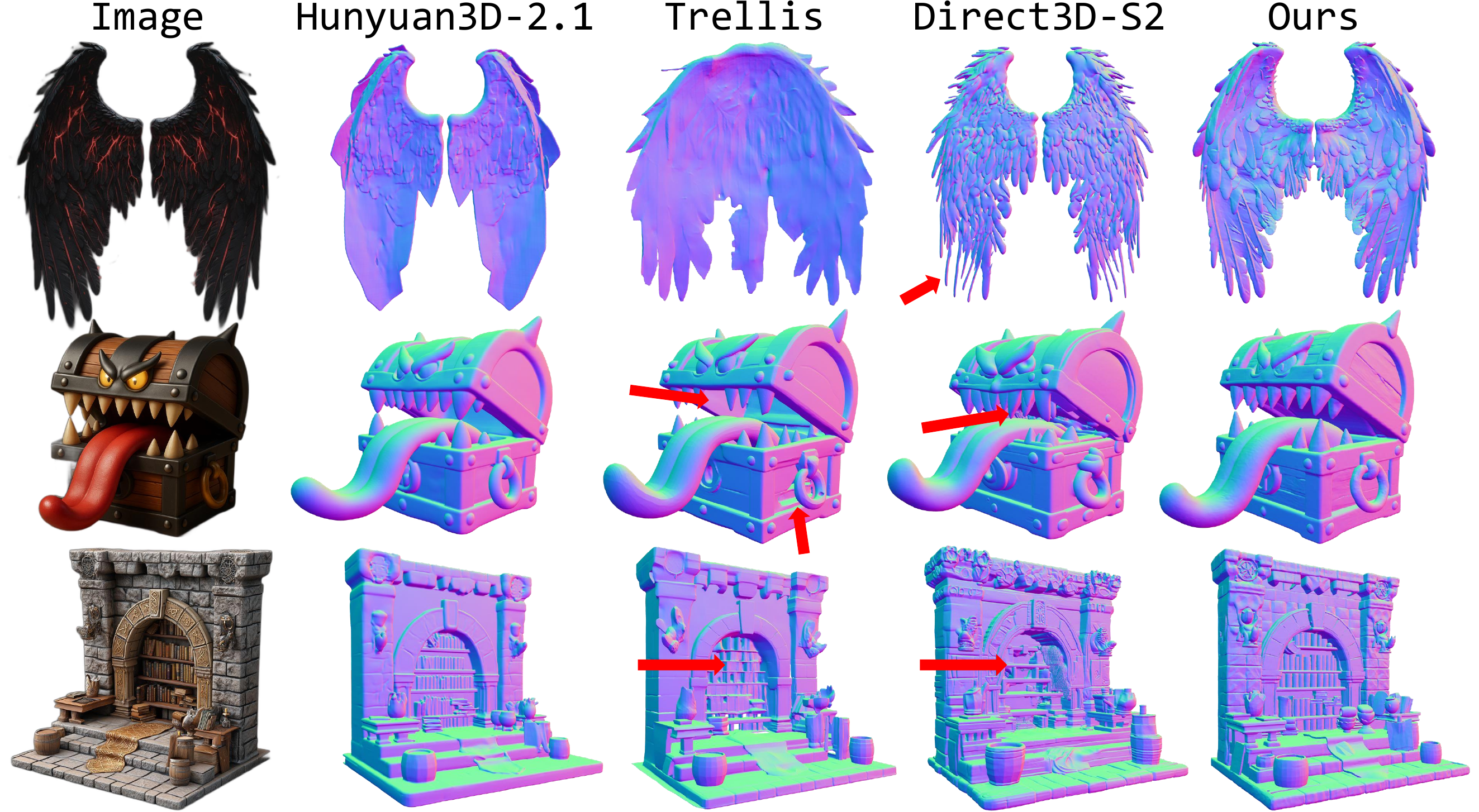}
    \vspace{-4mm}
    \caption{Visual Comparisons of image-to-3D generation.}
    \label{fig:generation}
    \vspace{-5mm}
\end{figure}

We validate the effectiveness of Topo-VAE as a powerful backbone for image-to-3D generation. 
We adopt the two-stage generation pipeline from Trellis~\cite{trellis}, but replace the structure generation stage with a VecSet-based model for stability, similar to Ultra3D~\cite{ultra3d}.
As shown in \cref{tab:generation} and \cref{fig:generation}, our results exhibit sharper geometric details and better alignment with input images, whereas other methods suffer from noise or holes (highlighted by red arrows).

\section{Limitation}
\label{sec:limitation}

Our sparse voxel-based VAEs generate millions of voxels when upsampling to high resolutions, which requires significant computational resources and time. Our remesh algorithm is limited by base resolution and therefore struggles to capture extremely fine details smaller than the voxel size.

\section{Conclusion}
\label{sec:conclusion}

In this paper, we present \ourmethod{}, a novel framework for high-fidelity mesh autoencoding. We identify the \textit{representation mismatch} as the key bottleneck in existing VAEs and propose the principle of Topological Unification, which enables direct supervision of fundamental mesh attributes. Extensive experiments demonstrate that our VAE achieves superior performance in reconstruction fidelity, especially in sharp features and fine geometric details.

\clearpage
{
    \small
    \bibliographystyle{ieeenat_fullname}
    \bibliography{main}
}

\end{document}